\documentclass{article} 
\usepackage{iclr2025_conference,times}

\usepackage{amsmath,amsfonts,bm}

\def\eqref#1{equation~\ref{#1}}

\def\1{\bm{1}}

\DeclareMathAlphabet{\mathsfit}{\encodingdefault}{\sfdefault}{m}{sl}
\SetMathAlphabet{\mathsfit}{bold}{\encodingdefault}{\sfdefault}{bx}{n}

\usepackage{hyperref}
\usepackage{url}
\usepackage{graphicx}
\usepackage{booktabs}
\usepackage{caption}
\usepackage{makecell} 
\usepackage{array}
\usepackage{booktabs}
\usepackage{amsmath} 
\usepackage{multirow}
\usepackage{listings}
\usepackage[utf8]{inputenc}
\title{SumForU: An LLM-based Review Summarization Framework
for Personalized Purchase Decision Support}

\author{
  Yuming Feng\thanks{Equal contribution.} \\
  Department of Electrical Engineering \\
  Stanford University \\
  Stanford, CA 94305, USA \\
  \texttt{yumingf@stanford.edu}
  \And
  Xinrui Jiang\footnotemark[1] \\
  Department of Electrical Engineering \\
  Stanford University \\
  Stanford, CA 94305, USA \\
  \texttt{jiangxr@stanford.edu}
}

\iclrfinalcopy 
\begin{document}
\maketitle

\begin{abstract}
Online product reviews contain rich but noisy signals that overwhelm users and hinder effective
decision-making. Existing LLM-based summarizers remain generic and fail to account for individual
preferences, limiting their practical utility. We propose \textbf{SUMFORU}, a steerable review
summarization framework that aligns outputs with explicit user personas to support personalized
purchase decisions. Our approach integrates a high-quality data pipeline built from the Amazon
2023 Review Dataset with a two-stage alignment procedure: (1) persona-aware Supervised Fine-Tuning (SFT)
via asymmetric knowledge distillation, and (2) Reinforcement Learning with AI Feedback (RLAIF)
using a preference estimator to capture fine-grained, persona-relevant signals. We evaluate the
model across rule-based, LLM-based, and human-centered metrics, demonstrating consistent
improvements in consistency, grounding, and preference alignment. Our framework
achieves the highest performance across all evaluation settings and generalizes effectively to unseen
product categories. Our results highlight the promise of steerable pluralistic alignment for building
next-generation personalized decision-support systems.
All code is publicly available at \url{https://github.com/Harry20030331/SumForU}.
\end{abstract}

\section{Introduction}

Current AI-powered review summarizers are often uniform, rigid, and saturated with impractical descriptions. They suffer from low informational density and utility. To address this, we reframe summarization as a \textbf{Human-Centered Alignment} problem, prioritizing diverse individual needs over a collective average.

We posit that a steerable pluralistic alignment paradigm is the most effective for this consumer decision-support task. This choice is justified as it avoids the inefficiency of presenting all possible views, which is an issue with Overton pluralism, and the high cognitive load of forcing users to interpret raw statistics, which is a drawback of distributional pluralism \citep{sorensen2024roadmap}. Our approach instead accepts an explicit user preference, such as a persona or query, to synthesize a customized, informative summary.

This leads to two core research questions:  
(1) \textbf{Mechanism Design:} How can we implement effective steering mechanisms, e.g., persona-aware Supervised Fine-Tuning (SFT) or reward modeling, that enable Large-Language Models (LLMs) to generate review summaries aligned with a user’s explicit persona?  
(2) \textbf{Effectiveness Evaluation:} To what extent does steerable summarization improve the summaries’ informativeness, factual reliability, and practical utility for decision-making?

To address these questions, we present a novel framework, \textbf{SUMFORU}, for personalized purchase decision support in this work. Our key contributions are as follows:
\begin{itemize}
    \item \textbf{High-Quality Data Pipeline:} We construct a rigorous dataset pipeline from the Amazon 2023 corpus, filtering for ``Active Users'' and ``Golden Products'' to ensure high signal-to-noise ratios. We introduce a pairwise construction method that pairs user personas with stratified review sets to simulate real-world decision scenarios.
    \item \textbf{Two-Stage Alignment Strategy:} We propose a hybrid training methodology combining Asymmetric Knowledge Distillation achieved via SFT, and Reinforcement Learning with AI Feedback (RLAIF). We demonstrate that while SFT provides stable behavior cloning, the subsequent RL stage is critical for capturing nuanced persona alignment.
    \item \textbf{Multi-Perspective Evaluation:} We validate our approach using a comprehensive triad of metrics, including Rule-based, LLM-based, and User-based, demonstrating that the high performance and generalization of our framework. 
\end{itemize}

\section{Related Work}
\textbf{Text Summarization.} The field of text summarization aims to distill large volumes of text into concise, informative representations. This domain has evolved significantly from early methods rooted in syntactic rules and statistical models to the current machine learning paradigm that directly learn from data \citep{SUPRIYONO2024100070}. In the scope of data-driven approach, this progression began with word embeddings such as Word2Vec \citep{mikolov2013efficient} that captures semantic relations and graph-based models identifying central sentences, and has now culminated in end-to-end neural networks and large-scale pre-trained models that understand deep, contextual patterns. This technological shift has enabled powerful, large-scale industrial applications, such as Apple's system for generating generic summaries from millions of App Store reviews, which leverages a multi-step information extraction form review process,  empowered by LLM with fine-tuned LoRA adapters \citep{hu2022lora} to ensure factuality and professionalism \citep{Apple2025ReviewSummarization}.

\textbf{Personalized and Steerable Summarization.} While powerful, the one-size-fits-all output of generic summarization models is ill-suited for personalization, as it fails to accommodate diverse user preferences and often overlooks critical, minority-held information. This limitation spurred research into personalization, especially in e-commerce application. Prior approaches improved salience estimation by incorporating user/product information and ratings \citep{xu2023pre}. This evolved into more sophisticated, persona-aware methods that treat user and product data as heterogeneous signals, such as separately modeling customer history for style and product history for common aspects to generate nuanced summaries \citep{cheng2023towards}. More recently, aspect-controlled models like MAPLE enabled explicit steering, but only via predefined categories \citep{yang2024maple}.

We build upon this progression to address a key gap: current models are either implicitly personalized to static history or explicitly steerable only by fixed categories. For our consumer decision-support task, we posit the steerable pluralistic alignment paradigm \citep{sorensen2024roadmap} is most effective. This approach avoids the inefficiency of Overton alignment that presents all views and the high cognitive load of Distributional alignment that forces users to interpret raw statistics). In contrast, our steerable model aligns with the user's core task by accepting their explicit preference, then synthesizing salient insights from reviews across all rating levels to provide a customized and informative summary for decision support.

\section{Data}\label{sec:data}

\begin{figure}[h]
\centering
\includegraphics[width=0.75\linewidth]{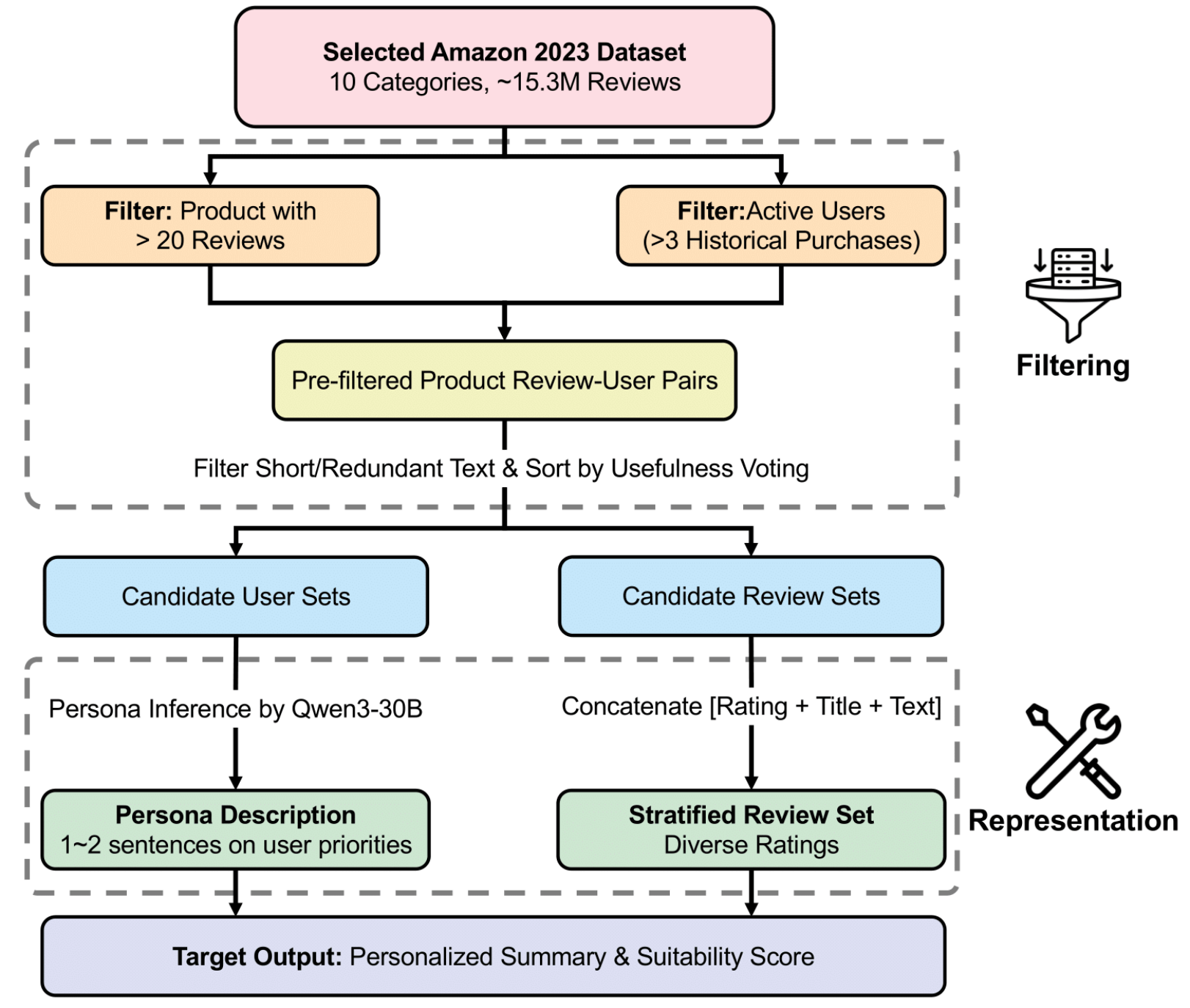}
\caption{Overview of our data pipeline.}
\label{fig:data_pipeline}
\end{figure}

Our framework is built on the Amazon 2023 Review Dataset~\citep{hou2024bridging}, which provides a massive corpus of 634{,}969 user reviews for 106{,}811 products written by 584{,}592 users. Each review contains a title, main text, rating, and helpful votes, and is traceable by user ID and timestamp. To construct a high-quality and task-relevant dataset, we followed a rigorous preprocessing and sampling pipeline, as illustrated in Fig.~\ref{fig:data_pipeline}.

\textbf{Identifying Active Users and Golden Products.}  
Since our study focuses on personalization, we first identified \textit{active users}, defined as users with at least three historical reviews, to provide a meaningful signal of preferences. Next, to ensure that our summarization task targets products with sufficient community discussion, we identified \textit{golden products}—products with at least 20 reviews and at least one review from an active user. This constraint filters out low-review products where users can easily read all comments, ensuring our task targets information-overloaded scenarios.

\textbf{Pairwise Construction and Standardization.}  
From the set of golden products, we constructed Product Review–User pairs by matching active users with the products they had reviewed. To simulate the information available at the time of purchase, we retained only reviews posted before the user’s own review timestamp. Each review set was then cleaned and standardized. We removed low-efficacy reviews with fewer than 5 words and filtered out positive reviews with ratings $>$ 3.0 but zero helpful votes, which often represent low-effort, low-information comments. 

To further standardize the input size, we discarded any product–user pair with fewer than 15 cleaned reviews. For overly large sets with more than 50 reviews, we applied stratified sampling based on the original rating distribution to maintain representativeness while normalizing the set to 50 reviews.

\textbf{Persona Construction.}  
For each active user, we generated a concise persona description using the \textbf{Qwen3-30B} model, derived from their past reviews. This persona serves as the personalization signal in our summarization framework.

\textbf{Final Dataset.}  
This pipeline yields high-quality pairs where each sample consists of an active user's persona and a stratified, standardized review set for a specific product. We use 300 pairs for training and 100 for evaluation within each product category, resulting in a total of 3{,}000 training and 1{,}000 testing pairs across different categories.

\section{Methodology}

\subsection{Overview}

The core design constraint of our methodology is computational efficiency, mandating an approach that is both effective and scalable . The framework thus utilizes a \textbf{lightweight language model}, specifically \textbf{Qwen3-4B-Instruct-2507}, selected for its optimal balance between reasoning capability and deployment feasibility.

\begin{figure}[h]
\centering
\includegraphics[width=0.9\linewidth]{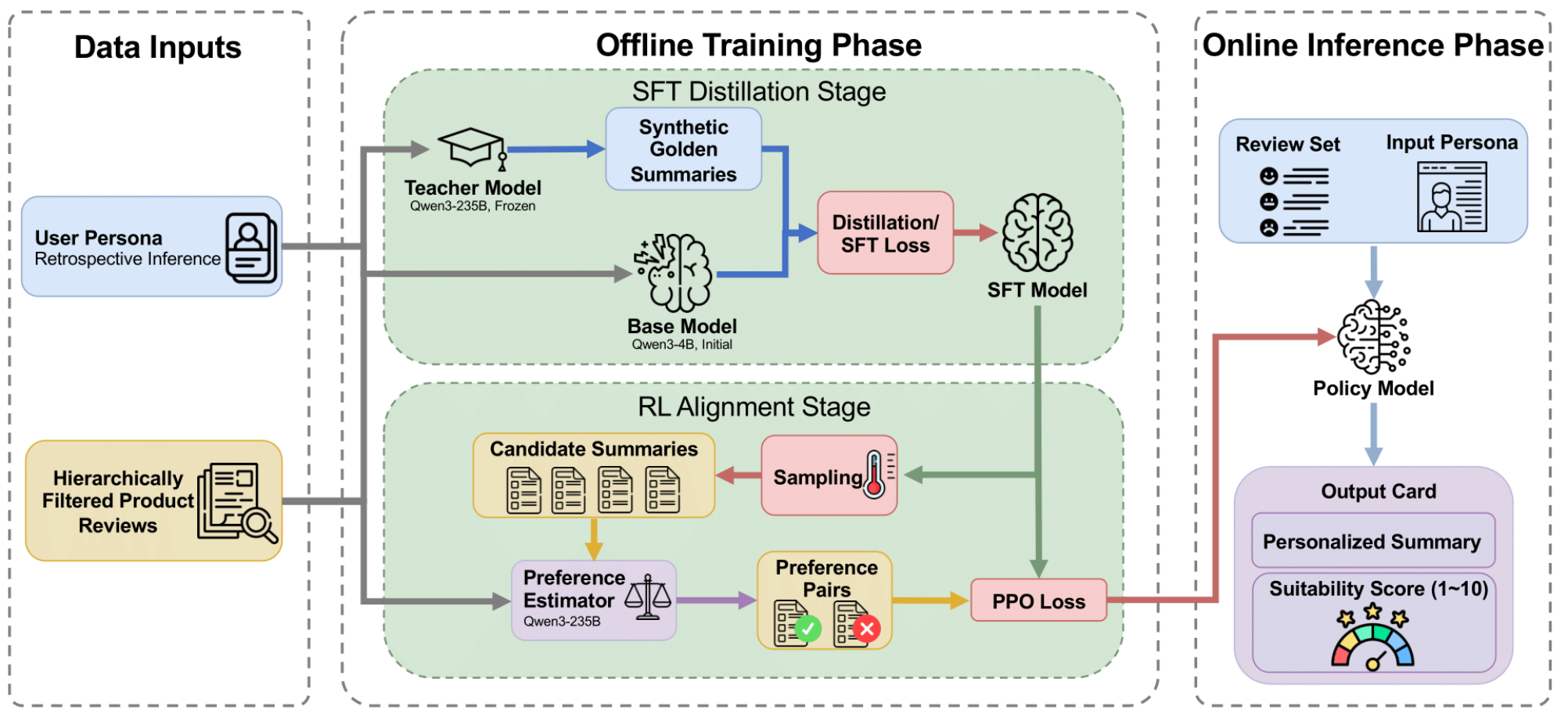}
\caption{Overview of SUMFORU pipeline.}
\label{fig:method_pipeline}
\end{figure}

Our process begins with two primary inputs: the \textbf{User Persona} and the \textbf{Review Set}, mentioned in Sec.~\ref{sec:data} . The goal is to produce a Dual Output Card, which includes a concise, \textbf{Personalized Summary} and a \textbf{Suitability Score} (1 to 10) . This score quantifies the product's alignment with the provided persona and offers a direct and quick suggestion for the user's purchase decision. The whole training and inference framework are shown in Fig.~\ref{fig:method_pipeline}.

\paragraph{Offline Training Phase}
The entire alignment system is structured through a two-stage offline training process designed to progressively refine personalization quality: \textbf{SFT}, shown in Sec.\ref{subsec:SFT}, and \textbf{RLAIF}~\cite{lee2023rlaif}, shown in Sec.\ref{subsec:RL}.

\paragraph{Online Inference Phase}
This phase defines the deployment scenario for the system. In the Online Inference Phase, a user directly provides their persona or specific personal demands, bypassing the LLM-based persona generation utilized during the training phase. The trained Policy Model immediately processes this live input and subsequently generates the dual-output card for user, which contains the summary and suitability score.

\subsection{Asymmetric Knowledge Distillation (Supervised Fine-Tuning, SFT)}\label{subsec:SFT}

To enhance model consistency and domain generalization, SFT was applied to the base model via asymmetric knowledge distillation using synthetic data generated by a larger teacher model. Specifically, the teacher model (\textbf{Qwen3-235B-A22B-Instruct-2507}) asynchronously produced high-quality Golden Summaries with sampling temperature $T=0.7$, which served as supervision signals. 

For each persona–reviews pair, the model input consists of the persona description and a collection of reviews, while the supervision target is the corresponding teacher-generated summary. Importantly, we intentionally excluded any human-written summaries of the target products to avoid teaching the model to imitate subjective and stylistic writing patterns. Instead, authentic user reviews of the product were included as part of the teacher model's input, providing an implicit but crucial alignment signal that helps the teacher better capture users' core priorities and decision factors, while this privileged information remains inaccessible to the student model.

SFT minimizes the cross-entropy loss ($\mathcal{L}_{\text{SFT}}$) to achieve behavior cloning, enabling the compact student model to approximate the teacher's conditional generation behavior. This stage establishes a stable initialization and injects essential alignment priors, forming the foundation for subsequent preference-based optimization.

\subsection{Reinforcement Learning with AI Feedback (RLAIF)}\label{subsec:RL}

While SFT enables the model to imitate the teacher's behavior, it does not guarantee that the generated summaries are optimal with respect to persona preference alignment. To address this gap, we apply a reinforcement learning (RL) procedure driven by AI-generated reward signals.

We begin by sampling multiple candidate summaries at a relatively high temperature to encourage stylistic diversity. These candidates are then evaluated by an AI-based \textbf{Preference Estimator}, implemented using \textbf{Qwen3-235B-A22B-Instruct-2507} and equipped with a tailored evaluation prompt (see Appendix~\ref{app:prompts}). The evaluator performs pairwise comparisons among candidates and assigns reward scores that reflect how well each summary aligns with the persona’s values.

Using these reward scores, we optimize the model via the Proximal Policy Optimization (PPO) objective, increasing the likelihood of generating high-reward summaries. This stage fine-tunes the SFT-initialized model using LoRA, enabling it to better capture persona-relevant cues and evidence prioritization, the capabilities that SFT alone cannot fully achieve.

\section{Results}

\subsection{Experimental Setup}
We conduct SFT and RL using the \textbf{Tinker Cookbook} framework, which provides high-level orchestration for model training and evaluation on top of PyTorch~\citep{paszke2019pytorch} and {Hugging Face Transformers library~\citep{wolf2019huggingface}. All models are trained with CUDA acceleration when available, automatically allocated by Tinker framework.

Training configurations are defined through Tinker's blueprint system, enabling reproducible runs with explicit hyperparameter control. The training hyperparameters and prompts can be found in Appendix.\ref{app:para} and \ref{app:prompts}.

\subsection{Comparison Analysis}\label{subsec:comparison_analysis}

This section presents a comparative analysis of four methods—Base Qwen-4B Model (\textbf{Base}), Instruction Prompt Tuning Model (\textbf{IPT}), \textbf{SFT}, and \textbf{RL}, across rule-based, LLM-based, and user-based evaluation perspectives.

\subsection*{Rule-based Metrics}
The rule-based evaluation uses objective, quantifiable metrics to assess two main aspects of the model outputs: Summary Text and Suitability Score.

\textbf{Summary Text Metrics}:
These metrics quantify \textbf{semantic similiarity} in the embedding space. \texttt{RefBS-R} (Reference BertScore Recall) measures semantic recall against the reference reviews. \texttt{RevBS-P} (Review BertScore Precision) assesses precision against the input reviews. \texttt{PersBS-R} (Persona BertScore Recall) evaluates recall for persona alignment information.

\textbf{Suitability Score Metrics}:
These metrics evaluate the \textbf{accuracy of the predicted preference score}. \texttt{MAE} (Mean Absolute Error) measures prediction error. \texttt{Spearman} assesses ranking consistency between predicted and true scores. \texttt{Within1Acc} (Within 1 Accuracy) evaluates the proportion of predictions with near-exact correctness ($\pm 1$ margin).

The results are shown in Table.~\ref{tab:rule_based}.
\begin{table}[h]
\centering
\caption{Rule-based Metrics: Summary Text and Suitability Score Evaluation}
\vspace{-3mm}
\label{tab:rule_based}

\setlength{\tabcolsep}{4pt}
\renewcommand{\arraystretch}{0.9}

\begin{tabular}{l c c c c c c}
\toprule
& \multicolumn{3}{c}{\textbf{Summary Text Metrics}} 
& \multicolumn{3}{c}{\textbf{Suitability Score Metrics}} \\
\cmidrule(lr){2-4} \cmidrule(lr){5-7}
\textbf{Method} & RefBS-R$\uparrow$ & RevBS-P$\uparrow$ & PersBS-R$\uparrow$ 
& MAE$\downarrow$ & Spearman$\uparrow$ & Within1Acc$\uparrow$ \\
\midrule
Base  & 0.7135 & 0.7618 & 0.8257 & 1.2362 & 0.4233 & 0.7007 \\
IPT   & 0.7146 & 0.7606 & 0.8172 & 1.2515 & 0.4498 & 0.6890 \\
SFT   & 0.7185 & 0.7542 & 0.8238 & 1.1130 & 0.5583 & 0.7720 \\
RL    & 0.7220 & 0.7516 & 0.8345 & 1.0780 & 0.5629 & 0.7640 \\
\bottomrule
\end{tabular}
\end{table}

\vspace{-3mm}

\textbf{Discussion.} 
Suitability score prediction metrics exhibit a clear upward trend (\texttt{Base} $\rightarrow$ \texttt{RL}), indicating the improvement ability of our model to capture user attitudes. However, traditional semantic metrics show minor differences, suggesting saturation on shallow semantics. Therefore, achieving effective \textbf{Persona Alignment} requires the use of \textbf{LLM Judges} to assess deeper semantic content and subtle contextual nuances.

\subsection*{LLM-based Metrics}

The LLM-based evaluation employs a large, expert Language Model to conduct a comparative analysis of generated summaries. This evaluation is performed by pairing summaries and scoring them across three predefined, critical dimensions: \textbf{Consistency} (score–text alignment), \textbf{Grounding} (faithfulness to reviews), and \textbf{Persona} (alignment to persona priorities), as shown in Table.~\ref{tab:llm_based}.

\begin{table}[h]
\centering
\caption{LLM-based Evaluation Results by Judge}
\vspace{-3mm}
\label{tab:llm_based}

\setlength{\tabcolsep}{2pt}
\renewcommand{\arraystretch}{0.85}

\begin{minipage}{0.47\linewidth}
\centering
\begin{tabular}{l c c c c}
\toprule
\multicolumn{5}{c}{\textbf{Judge: Qwen3-235B}} \\
\midrule
Method & Overall$\uparrow$ & Consis.$\uparrow$ & Ground.$\uparrow$ & Persona$\uparrow$ \\
\midrule
Base & 0.336 & 0.441 & 0.331 & 0.235 \\
IPT  & 0.489 & 0.435 & 0.709 & 0.322 \\
SFT  & 0.507 & 0.440 & 0.531 & 0.551 \\
RL   & 0.668 & 0.684 & 0.429 & 0.892 \\
\bottomrule
\end{tabular}
\end{minipage}
\hfill
\begin{minipage}{0.47\linewidth}
\centering
\begin{tabular}{l c c c c}
\toprule
\multicolumn{5}{c}{\textbf{Judge: gpt-oss-120b}} \\
\midrule
Method & Overall$\uparrow$ & Consis.$\uparrow$ & Ground.$\uparrow$ & Persona$\uparrow$ \\
\midrule
Base & 0.441 & 0.464 & 0.361 & 0.497 \\
IPT  & 0.439 & 0.419 & 0.466 & 0.431 \\
SFT  & 0.490 & 0.488 & 0.532 & 0.452 \\
RL   & 0.630 & 0.629 & 0.642 & 0.620 \\
\bottomrule
\end{tabular}
\end{minipage}
\end{table}

\textbf{Discussion.} 
Both judges consistently rank RL as the top method.
While Qwen shows a stronger RL advantage potentially due to reward-model similarity, GPT results confirm the gain is genuine.
These results highlight RL's ability to improve persona alignment, consistency, and factual grounding beyond what SFT or prompt tuning can achieve. However, a key concern is the \textbf{circular dependency} created by using LLM-trained networks evaluated by LLM judges. This overlap in capabilities between training and assessment may amplify model biases, underscoring the necessity of \textbf{User-based Metrics} for a more objective evaluation.

\subsection*{User-based Metrics}
To counteract the potential biases of the LLM-based evaluation and provide a definitive measure of human preference, we conducted a small-scale user study.

Three annotators evaluated 10 distinct cases, sampling one from each category. The evaluation process involved two stages: First, each annotator read the provided persona and review set, then ranked the summaries generated by the four methods based on their perceived \textbf{usefulness and helpfulness}. Second, for the top-ranked summary, annotators performed a \textbf{Targeted Likert Validation} using a 5-Point Likert Scale (1 = Strongly Disagree, 5 = Strongly Agree) across three key dimensions: Persona Alignment, Decision Utility, and Factual Trustworthiness. 

This dual-stage methodology ensures data collection efficiency by strategically focusing human effort solely on the most valuable output. While this user study was necessarily small-scale due to resource constraints, it nevertheless provides real, human-centered insight into user preferences.

The results of user-based metrics are shown in Table ~\ref{tab:user_ranking}
 and Table ~\ref{tab:rl_likert}.
 
\begin{table}[h]
\centering
\caption{User-based Ranking Metrics}
\vspace{-3mm}
\label{tab:user_ranking}

\setlength{\tabcolsep}{4pt}
\renewcommand{\arraystretch}{0.9}

\begin{tabular}{l c c c c}
\toprule
\textbf{Metric} & \textbf{Base} & \textbf{IPT} & \textbf{SFT} & \textbf{RL} \\
\midrule
Win Rate-LLM       & 0.1 & 0.1 & 0.0 & 0.8 \\
Win Rate-Humans    & 0.0 & 0.1 & 0.1 & 0.8 \\
Mean Rank-LLM      & 3.3 & 2.8 & 2.7 & 1.2 \\
Mean Rank-Humans   & 3.8 & 2.8 & 2.2 & 1.2 \\
\bottomrule
\end{tabular}
\end{table}

\vspace{-2mm}

\begin{table}[h]
\centering
\caption{Targeted Likert Validation Scores for RL Method}
\vspace{-3mm}
\label{tab:rl_likert}

\setlength{\tabcolsep}{6pt}
\renewcommand{\arraystretch}{0.9}

\begin{tabular}{c c c}
\toprule
\textbf{Persona Alignment} & \textbf{Decision Utility} & \textbf{Factual Trustworthiness} \\
\midrule
4.875 & 4.917 & 4.792 \\
\bottomrule
\end{tabular}
\end{table}

\textbf{Discussion.} The User-based Metrics provide critical human validation for our findings. First, the high annotator agreement (Kendall's $W = 0.787$) confirms the reliability of the human rankings. Second, the RL model achieved the highest win rate and best mean ranking, directly mirroring the LLM-judge results. This congruence validates the use of the LLM-judge as an efficient proxy for human preference. Finally, the Likert validation confirms the absolute quality of RL summaries. This collectively demonstrates RL's effectiveness in optimizing summaries against complex, human-centric criteria.

\subsection{Category-wise Analysis}\label{subsec:category_analysis}

To gain insight into how different product categories influence the effectiveness of our RL framework, we conducted a detailed, category-wise analysis. This process is essential for identifying category-specific strengths and weaknesses in RL optimization, enabling more targeted model improvements and ensuring robustness across categories. The results are demonstrated in Fig.~\ref{fig:category_result}.

\begin{figure}[h]
\centering
\includegraphics[width=1.0\linewidth]{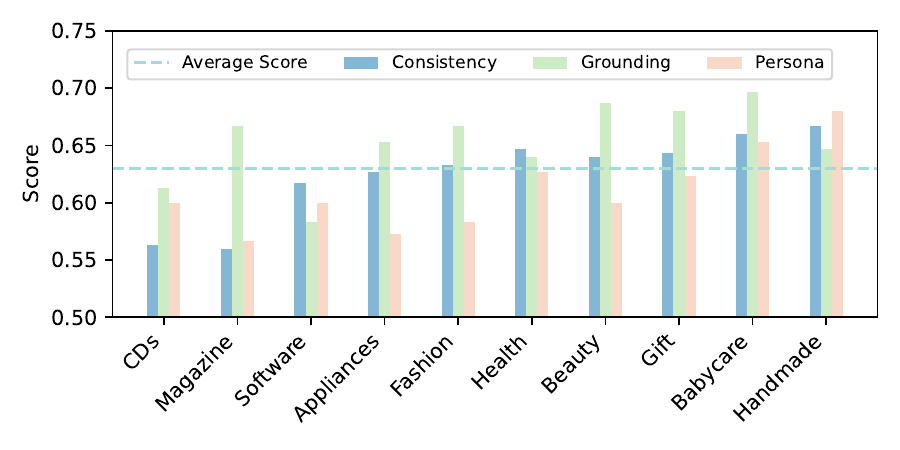}
\caption{RL method performance across product categories.}
\label{fig:category_result}
\end{figure}

\vspace{-1mm}
\paragraph{Observed metric gap.}
CDs shows substantially worse performance than Handmade on all three subjective metrics: Consistency (CDs: 0.563 vs Handmade: 0.667), Grounding (CDs: 0.613 vs Handmade: 0.647), and Persona (CDs: 0.600 vs Handmade: 0.680). Because these gaps are systematic, we investigate underlying data properties to determine whether they explain the discrepancy and how they might explain how our framework works.
  \vspace{-1mm}
\paragraph{Data comparison (key statistics).}
A brief validation of the underlying data reveals the following:  \vspace{-6mm}

\begin{enumerate}
  \item The two categories have nearly identical average reviews per entry (difference $<$ 1\%).
  \vspace{-1mm}
  \item Their overall mean ratings are similar (relative difference $\approx$ 5\%).
  \vspace{-1mm}
  \item The most prominent difference is in per-product rating variance (Avg Var/Product):
    \vspace{-1mm}
  \[
    \text{CDs: } 1.1718 \quad\text{vs.}\quad \text{Handmade: } 1.6971,
  \]
    
  which corresponds to a $\approx$44.8\% higher Avg Var/Product for Handmade.
\end{enumerate}

\vspace{-2mm}
\paragraph{Interpretation and implications.}
Because data volume and mean-rating are comparable, the key distinguishing factor is \textbf{Arg Var/Product}. Handmade's higher per-product variance implies more diverse and informative user feedback, which provides richer training signals that RL can exploit. CDs, by contrast, exhibits lower variance and thus a relatively homogeneous feedback signal, which can be handled by Base Model very well, limiting RL performance gains.

\subsection{Generalization Analysis}

To assess the generalization capability of our RL-trained model, we evaluated its performance on categories excluded from the training set. This critical check addresses potential overfitting, given our relatively small training dataset ($\sim$3k). The model's real-world utility hinges on its ability to generalize effectively across a vast array of unseen product domains on e-commerce platforms.

We directly deployed the RL-trained model on a new test set consisting of three unseen categories: \textbf{Video\_Games}, \textbf{Arts\_Crafts\_and\_Sewing}, and \textbf{Industrial\_and\_Scientific}. Each category contained 100 data entries. We evaluated performance using Ruled-based and LLM-based metrics, comparing RL against the Base to quantify generalization improvements.
The results can be found in Table.~\ref{tab:generalization_full_metrics}.

\vspace{-2mm}
\begin{table}[h]
\centering
\caption{Generalization Results on Unseen Categories (Full Metrics)}
\vspace{-3mm}
\label{tab:generalization_full_metrics}
\begin{tabular}{l c c c c c c}
\toprule
\textbf{Category} & \textbf{Method} & \textbf{MAE} $\downarrow$ & \textbf{Spearman} $\uparrow$ & \textbf{Consis.} $\uparrow$ & \textbf{Ground.} $\uparrow$ & \textbf{Persona} $\uparrow$ \\
\midrule
\multirow{2}{*}{VideoGames}
    & Base & 1.1300 & 0.5199 & 0.443 & 0.360 & 0.440 \\
    & RL & 1.0500 & 0.5522 & 0.630 & 0.630 & 0.637 \\
\midrule
\multirow{2}{*}{Arts Crafts}
    & Base & 1.3550 & 0.3184 & 0.480 & 0.390 & 0.457 \\
    & RL & 1.1600 & 0.5335 & 0.640 & 0.587 & 0.617 \\
\midrule
\multirow{2}{*}{Industrial Product}
    & Base & 1.3100 & 0.4882 & 0.500 & 0.310 & 0.463 \\
    & RL & 1.0600 & 0.5418 & 0.623 & 0.683 & 0.607 \\
\bottomrule
\end{tabular}
\end{table}

\vspace{-2mm}
\textbf{Discussion.} 
RL demonstrates improvements across all evaluation metrics, successfully proving the framework's capability to \textbf{generalize effectively to unseen categories}.
However, the MAE improvement in the Video Games category is relatively minor (an improvement of 0.0800), representing only about 50\% of the overall improvement observed in the comparison analysis (0.1582, shown in Sec.\ref{subsec:comparison_analysis}). This highlights variability in RL's generalization performance on untrained categories, suggesting that further optimization is still needed to enhance cross-category robustness.

Finally, to illustrate how our framework behaves in real-world decision scenarios, we provide a qualitative case study highlighting persona alignment, evidence grounding, and difference across methods (see Appendix~\ref{app:case_study}).

\section{Conclusion}
\vspace{-1mm}
In this work, we presented a steerable summarization framework that effectively aligns with explicit user preferences and latent attitudes. Our experimental results highlight several key findings:
\begin{enumerate}
\vspace{-2mm}
    \item \textbf{Superiority of Our Framework:} the \textbf{RL} method demonstrates the strongest alignment with user persona and best decision utility across rule-based, LLM-based, and user-based metrics.

    \item \textbf{Effectiveness in Bipolar Situations:} Our category-wise analysis revealed a specific strength of the \texttt{RL} model in capturing and summarizing bipolar opinions, ensuring the resulting summaries acknowledge and integrate diverse, conflicting feedback more reasonably than the \texttt{Base} model.

    \item \textbf{Generalization and Data Efficiency:} Our RL-trained model successfully generalized to unseen categories, demonstrating the effectiveness and strong generalization capability achieved through fine-tuning on a relatively small preference dataset.
\vspace{-2mm}
\end{enumerate}

Future work will focus on two primary directions: enhancing cross-category robustness to improve zero-shot generalization further, and optimizing deployment and inference efficiency to create a real and seamless user experience. Ultimately, this framework establishes a robust baseline for the next generation of personalized, agentic decision-support systems.

\section*{Acknowledgments}
We would like to express our sincere gratitude to Prof. Diyi Yang for her invaluable guidance and for organizing this insightful course. We also extend our special thanks to the teaching staff, specifically Sunny Yu and Advit Deepak, for their constructive feedback, technical advice, and helpful discussions throughout the development of this project.

\newpage

\section*{Ethical Consideration}
\textbf{Data Privacy and User Autonomy.} Our research utilizes the publicly available Amazon 2023 Review Dataset, mitigating risks associated with unauthorized data collection. Crucially, we distinguish between the retrospective analysis used for training and the privacy-preserving nature of deployment. Unlike systems that harvest behavioral logs, the deployed SUMFORU operates exclusively on transient, real-time user inputs. By decoupling the service from long-term history, we eliminate the need for persistent profiling, ensuring that the persona remains a temporary instruction controlled entirely by the user.

\textbf{Independence and Mitigation of Conflict of Interest.} To prevent goal misalignment, where platforms might manipulate models to prioritize sales over user satisfaction—we advocate for deploying SUMFORU as an independent, third-party user agent, e.g., a browser extension. Decoupling the summarization utility from the marketplace vendor allows the framework to act as an objective auditor. This independence ensures that the Suitability Score  remains faithful to the user's explicit needs, free from commercial optimization pressures or biases that often plague platform-native algorithms.

\textbf{Safety in High-Stakes Domains.} Domains For real-world deployment, particularly when extending to sensitive domains such as accommodation or travel, we advocate for the implementation of a Hybrid Safety Protocol. This approach augments the generative model with deterministic, rule-based filters designed to identify non-negotiable safety keywords. These alerts are engineered to bypass personalization settings and are forcibly presented to the user, ensuring that the pursuit of a tailored experience never suppresses critical warnings or compromises physical safety.

\section*{Author Contributions}
\vspace{-1mm}
\textbf{Yuming Feng} and \textbf{Xinrui Jiang} contributed equally to this work.
\begin{itemize}
    \vspace{-1mm}
    \item \textbf{Yuming Feng}: implemented the SFT and RL training frameworks, developed the automated LLM-based evaluation, conducted the category-specific and generalization experiments, performed the comprehensive data analysis.
    \item \textbf{Xinrui Jiang}: conceived the original project vision, designed and implemented the fine-grained data processing pipeline, developed the rule-based evaluation, designed and conducted the user study protocols.

    \vspace{-1mm}
\end{itemize}
Both authors collaborated closely on the system design and the final manuscript preparation.

\section*{Appendix}
\appendix

\section{Case Study}\label{app:case_study}
This case study assesses the product fit of a synthetic lace-front wig for customers demanding realism and styling control. Sections below provide the persona, product reviews, and an objective assessment of different model summaries.

\subsection{Persona}
This customer prioritizes \textbf{realistic appearance} and \textbf{styling flexibility} in wigs, and is dissatisfied with products that lack \textbf{customization options} or reveal \textbf{visible construction details}, indicating a preference for \textbf{high-quality}, \textbf{natural-looking} wigs with \textbf{versatile styling potential}.

\subsection{Selected Reviews}

\begin{itemize}
    \item \textbf{Review 1: rating 5.0. Beautiful Wig!} (especially for the price) This wig is DEF a buy! It's a light Swiss lace with a large amount of parting space on the crown. Several inches of ventilation on the sides with varied tying to create a \textbf{more natural hairline}. Pre-cut baby hairs at the nape of the neck! Super soft! Doesn't seem to tangle easily. It's very easy to comb my hands through. The color is a beautiful ashy blonde. A cool, 613 color, with nice dimension to it. Overall, a defi...

    \item \textbf{Review 2: rating 5.0. WOW!} Wow is all i can say! When you see wigs on here for such a good deal it sometimes seems too good to be true. I am completely in love with this wig! I am attaching a picture and a short video. The picture is after I \textbf{cut the lace}, put on a wig cap to cover my half black hair, used a small amount of hair glue and then \textbf{blended in with my foundation}. The video is before I finsihed cutting the lace and did all of the above. First, I am a HUG...

    \item \textbf{Review 3: rating 5.0. Great bang for your buck}; beautiful blonde shade. I will update after a few more wears, but so far, so good. I cut it, \textbf{did a curl set} and steamed the curls with a clothing steamer to set and brushed out. The shade of blonde is very pretty, it is sort of a lightest ash blonde, not entirely platinum but not as yellow as a \#613. It has dimension and just the right amount of shine. Super soft. It will obviously tangle, being so long and synthetic, but so far I wo...

    \hspace{10em}\textbf{\ldots{}(skip 6 similar rating 5.0 reviews)}

    \item \textbf{Review 10: rating 1.0. Do not buy!} If i could rate it zero stars i would! It was \textbf{choppy and uneven} not one length and \textbf{the tracks were showing}. And the lace was \textbf{dirty looking}. It was disgusting!

    \item \textbf{Review 11: rating 3.0. Damaged?} Prepurchased? Or just made from bad hair? I believe this wig would have been pretty damn good if it hadn't been \textbf{worn before me} or some how damaged before getting to me. I don't know what the case is I just know there were two spots inside the cap where the clips to adjust the cap are that were \textbf{visibly dirty}. And once the wig was on the entire top back half of the wig is \textbf{all broken} little hairs that stick straight up making the wig \textbf{look tore up} for being brand new. Also as so...

    \item \textbf{Review 12: rating 3.0. No parting space.} The wig is very soft, looks beautiful, but there is absolutely no parting space. Like absolutely none. You can \textbf{only wear it like in one style} that will hide the tracks. That cute little braid up style looks a mess in real life because all everyone can see is the weave tracks. Near the weave tracks the hair is not so silky. The wig would be perfect if it had some parting space and more lace in the front.
    
    \hspace{9em}\textbf{\ldots{}(skip 8 similar rating 1.0--3.0 reviews)}
    
\end{itemize}

\subsection{Model Output}
\subsubsection*{Base}
\quad Summary: This wig delivers a \textbf{highly realistic, natural appearance} with \textbf{ample parting space}, Swiss lace construction, and \textbf{versatile styling options} that blend seamlessly with the user's skin and hairline. It features soft, durable hair with authentic dimension and minimal shedding, making it ideal for daily wear and creative styling. 

\quad Suitability: 9/10

\subsubsection*{IPT}
\quad Summary: The wig delivers a \textbf{natural, realistic appearance} with \textbf{excellent styling flexibility}, featuring \textbf{ample parting space}, smooth hairline integration, and \textbf{versatile styling options} that \textbf{hide construction details}. High praise for softness, dimension, and \textbf{realistic look}, especially when \textbf{cut and styled properly}. 

\quad Suitability: 8/10 - Strong match for customers seeking \textbf{realistic styling} and \textbf{natural appearance}, though lower ratings stem from \textbf{poor quality} or \textbf{visible tracks} in subpar versions.

\subsubsection*{SFT}
\quad Summary: The wig receives strong praise for its \textbf{realistic appearance}, soft texture, and \textbf{natural look} when styled, with many noting its suitability for cosplay and everyday use; however, multiple reviews highlight issues with \textbf{visible lace tracks}, \textbf{lack of styling flexibility}, and \textbf{inconsistent quality control}, particularly for front lace wigs seeking \textbf{customization}. While several users report excellent results after styling and hairline adjustments, others found the construction too tight or damaged upon arrival, undermining the natural look and styling potential. 

\quad Suitability: 6/10 - Mixed feedback on \textbf{customization} and \textbf{construction quality} makes it a moderate fit for those prioritizing \textbf{realistic, flexible styling}.

\subsubsection*{RL}
\quad Summary: The wig receives widespread praise for its \textbf{realistic appearance}, softness, and \textbf{natural hairline} with ventilation and \textbf{parting space} that supports \textbf{styling flexibility}, though multiple low ratings highlight \textbf{serious inconsistencies in quality}, including \textbf{visible lace tracks}, shedding, tangling, and poor fit, which directly conflict with the persona's need for \textbf{customization} and \textbf{undetectable construction details}. Despite several users reporting excellent results and high satisfaction, the repeated complaints about \textbf{authenticity}, durability, and mismatched visuals indicate a product that fails to meet expectations for \textbf{high-end, natural-looking, and fully style-able wigs}. 

\quad Suitability: 4/10 - \textbf{Poor consistency} and frequent reports of \textbf{visible construction flaws} make it unsuitable for a customer prioritizing \textbf{realistic appearance} and \textbf{styling customization}.

\subsection{Analysis}
The wig reviews show bipolar differentiation: high praise for natural aesthetics versus criticism of quality flaws. The Base model consistently ignores negative feedback, while the RL model sharply captures adverse details and makes reasonable judgments, highlighting its superior sensitivity to inconsistencies, which we also found before in Sec. \ref{subsec:category_analysis}.

\section{Hyperparameters}
\label{app:para}
We utilized the \texttt{Tinker Cookbook} framework for all experiments. The hyperparameters for the Supervised Fine-Tuning (SFT) and Reinforcement Learning (RL) stages are detailed in Table ~\ref{tab:hyperparameters}.
\vspace{-2mm}
\begin{table}[h]
\centering
\small
\caption{Hyperparameters for Qwen3-4B across SFT and RL stages.}
\vspace{-3mm}
\label{tab:hyperparameters}
\begin{tabular}{lcc}
\toprule
\textbf{Hyperparameter} & \textbf{SFT Distillation} & \textbf{RL Alignment} \\
\midrule
Learning rate & 2e-4 & 1e-5 \\
Batch size & 16 & 16 \\
Epochs & 10 & 1 \\
Max sequence length & 8192 & 8192 \\
\midrule
Optimizer & Adam & Adam \\
Adam $\beta_1$ & 0.9 & 0.9 \\
Adam $\beta_2$ & 0.95 & 0.95 \\
Adam $\epsilon$ & 1e-8 & 1e-8 \\
\midrule
LoRA rank ($r$) & 32 & 32 \\
\bottomrule
\end{tabular}
\end{table}

\section{Prompts}
\label{app:prompts}

Below are the core prompts used in our pipeline, formatted for the system and user roles.
\vspace{-3mm}
\subsection{Persona Inference Prompt}
\begin{lstlisting}[basicstyle=\ttfamily\scriptsize, breaklines=true, frame=single, numbers=none]
########################
# System Prompt
########################
You are an expert market analyst specializing in customer personas.
Generate the persona text directly, without any prefix or explanation.
Make the persona concise and focused on key traits, in ONLY ONE sentence.

Below are examples of how to generate customer personas based on product reviews.
Example 1:
Reviews:
-- "The product arrived quickly and was packaged securely."
-- "Very satisfied with the quality, feels durable and well-made."
Persona:
This customer values product quality and reliable delivery speed. They are less sensitive to price and prioritize a smooth, trustworthy shopping experience.
Example 2:
Reviews:
-- "It's affordable and works well for daily use."
-- "Good deal for the price, I'll buy again during discounts."
Persona:
This customer is price-conscious and tends to look for good value and deals. They enjoy functional products that balance cost and performance.

########################
# User Prompt
########################
Now, based on the following product reviews, create a concise persona of the typical customer:

{combined_reviews}

Provide insights into their purchase preferences.
Please don't generate any persona descriptions not related to purchase preferences, like age, gender, or privacy. Generate the persona text directly, without any prefix, label, or explanation.
\end{lstlisting}

\subsection{Policy Model Prompt (Summary Generation)}
\begin{lstlisting}[basicstyle=\ttfamily\scriptsize, breaklines=true, frame=single, numbers=none]
########################
# System Prompt
########################
You are an intelligent review summarization assistant.
Your task is to read a set of user reviews and generate a concise, decision-oriented summary that reflects what the target persona values most, while identifying consistently mentioned weaknesses that could affect satisfaction.

Instructions:
1. Goal: Generate a 2~3 sentence summary that helps the persona quickly decide whether the product fits their needs.
2. Tone: Neutral but empathetic (objective and informative, not promotional). Avoid repetitive listing; merge overlapping opinions naturally.
3. Reasoning Process
    Follow this structured reasoning before producing the final output:
    Step 1. **Interpret Persona:** Identify the persona's core priorities and evaluation criteria (e.g., durability, price, comfort). Determine what the persona values most and what factors matter for their satisfaction.
    Step 2. **Analyze Reviews:** Detect recurring patterns across rating levels while filtering out noise and isolated opinions. Focus on themes, strengths, and weaknesses mentioned multiple times, and ignore isolated or contradictory negatives that appear only once.
    Step 3. **Integrate Evidence:** Synthesize representative insights that reflect both positive and negative aspects relevant to the persona. Ensure that these insights align with the persona's values and are grounded in multiple review observations.
    Step 4. **Write Summary:** Compose a 2~3 sentence personalized summary that integrates persona priorities with aggregated user perspectives. Include a **1~10 suitability rating** with a concise justification referencing consistent evidence from the reviews.
4. Content requirements: Be concise. Do not invent information outside the reviews. Suitability rating reflects the overall match between the persona's expectations and review sentiment(1~10).

Few-shot Examples:
Case 1 -- Not Recommended
<Input>
persona: A practical user who values product durability and quality over price.
reviews:
-- The bottle leaked after a week, not worth it.
-- Works okay at first, but broke down quickly.
-- Packaging feels cheap and the lotion smells odd.
<Output>
Summary: The product shows inconsistent quality and poor durability, making it unreliable for long-term use.
Suitability: 3/10 -- Not recommended for users prioritizing lasting quality.

Case 2 -- Acceptable
<Input>
persona: A casual buyer who values decent performance at a fair price.
reviews:
-- Does what it says, though not perfect.
-- Feels okay on skin, but a bit greasy.
-- Good for the price, but could absorb faster.
<Output>
Summary: This product offers reasonable value and moderate performance, suitable for those with basic expectations.
Suitability: 6/10 -- Acceptable choice for budget-conscious users.

Case 3 -- Highly Recommended
<Input>
persona: A skincare enthusiast seeking visible anti-aging results.
reviews:
-- My skin feels smoother and looks brighter.
-- Great texture and easy to apply.
-- I've received compliments after just two weeks!
<Output>
Summary: The product delivers visible rejuvenation and enhances skin texture, showing clear anti-aging benefits.
Suitability: 9/10 -- Highly recommended for users seeking effective skincare results.

Important:
- You MUST produce both "Summary:" and "Suitability:" exactly as shown.
- Start your output with "Summary:" and nothing else.
- Do NOT include any introduction or extra text.
- Example of the correct output format:
Summary: <concise description of the product's strengths, performance, and consistent issues if repeatedly mentioned>
Suitability: <1~10 rating> -- <short justification in one phrase>

########################
# User Prompt
########################
Help write a personalized product summary for a customer based on their purchase persona from product reviews set.
Output a CONCISE summary in 2~3 sentences(Summary: ) and 1~10 suitability rating(e.g. Suitability: 8/10), without any prefixes and extra explanations.

Input Format:
persona: <persona>
product reviews: <reviews>
\end{lstlisting}

\bibliography{refs}
\bibliographystyle{iclr2025_conference}

\end{document}